# From Manual Construction to AI-Driven Scenario Emergence: Rethinking Catastrophe Risk Modeling

by Hang Gao

## Abstract

Traditional catastrophe (CAT) risk models rely on costly manual construction to generate extreme weather scenarios, an approach largely unchanged since the 1990s. As climate extremes intensify, this creates mounting challenges to the entire risk transfer chain. This study proposes the TAISE framework, which repurposes AI weather forecasting models to produce coherent extreme weather sequences at a fraction of traditional costs. Through self-iterative generation, the framework produces continuous global atmospheric fields from which extreme events emerge. A proof-of-concept experiment demonstrates an order-of-magnitude reduction in computational cost compared with conventional methods, while capturing temporal continuity and cross-regional correlations absent in snapshot-based approaches. These findings suggest a pathway toward democratising catastrophe risk quantification and enabling dynamic, comprehensive portfolio assessment for insurers, reinsurers, ILS fund managers and public-sector risk managers.

**Keywords:** Catastrophe risk modeling, Artificial intelligence (AI) in insurance, Climate change risk, Insurance technology

# 1. Introduction

## 1.1 Research Background

The insurance industry is fundamentally dependent on CAT models for risk transfer and portfolio management. As climate change significantly amplifies the frequency and intensity of global extreme weather events, the long-term upward trend in catastrophe-related insured losses has become increasingly pronounced (Kemp et al., 2022; Yuryeva, Kovaleva and Shukhova, 2023; Davidson and Kemp, 2024). This trajectory has exposed the structural deficiencies of traditional CAT models and placed mounting pressure on the entire risk transfer value chain, from primary insurers and reinsurers, to insurance-linked securities (ILS) investors and structurers alike.

The sluggish pace of model updates makes it difficult for the insurance industry to flexibly reflect continuous shifts in climate baselines in a timely manner (Bahl et al., 2024). Most critically, the prohibitive development costs and low efficiency of traditional methodologies have severely restricted their widespread application. Despite nearly four decades of industry use, a recent market research survey by Aon found that, 48% of insurance companies have not authorized the use of CAT models, and only 27% of institutions maintain dedicated teams for model evaluation (Aon plc, 2025). Furthermore, for entities tasked with social risk management— such as multilateral development banks, policy-oriented financial institutions, and NGOs—budget constraints often render these essential risk quantification tools inaccessible (Berger, Emmerling and Tavoni, 2017; Mignan, 2024; Higuera Roa et al., 2025).

The explosive growth of AI and high-performance computing (HPC) is exerting an unprecedented impact across numerous sectors, including financial services (Ashta and Herrmann, 2021; Lodhi, Gill and Hussain, 2024). Business operations previously characterized by high costs and complex workflows are now witnessing a significant reduction in operational expenses. This technological disruption is dissolving traditional market entry barriers and reshaping competitive dynamics across industry. While the strategic shift toward large language models (LLMs) and automated coding has already triggered structural change within broader financial services (Ashta and Herrmann, 2021; Wosu et al., 2025), the application of AI-native methods to CAT modeling remains largely unexplored.

In recent years, AI has demonstrated remarkable capabilities in the meteorological domain (Harris et al., 2022; Koldunov et al., 2024; Lodhi, Gill and Hussain, 2024; Kow et al., 2025; Manco, 2025). Nevertheless, the insurance sector's response to this technological wave has been lagging. Current explorations are largely confined to incremental technical refinements within the model development process – efforts that have yet to address the core of catastrophe risk modeling: the generation paradigm of Stochastic Event Sets (*SES*).

This study conducts a critical assessment of traditional CAT modeling approaches and investigates the feasibility of transitioning from traditional 'manually-led extrapolation' to 'AI-driven scenario emergence'. To this end, we propose the Temporal AI-Based Stochastic Event generation Framework (TAISE Framework, and present a preliminary feasibility test of its core component, the Genesis Module.

### 1.2 Structural Limitations of Traditional Methodologies

Professional CAT models were standardized in the 1990s. Their core logic involves generating SES to compensate for the scarcity of historical catastrophe samples, thereby satisfying the Law of Large Numbers required for actuarial quantification (Cobb, 1978; Mignan, 2024).

A comprehensive CAT model typically comprises three modules: the Hazard Module, which generates SES; the Vulnerability Module, which translates physical hazard intensity into asset damage; and the Financial Module, which converts physical damage into financial loss and risk metrics. The model produces Exceedance Probability (EP) curves to support a suite of decision-making needs, including pricing, treaty structuring, portfolio management and so on. The Hazard Module (SES generation) serves as the foundation of the entire model; it is also the most capital-intensive component and the stage where manual labor are most concentrated.

Constrained by the limited computing power of the PC era, the traditional SES generation method relied heavily on manually-led extrapolation. This involves parametric cloning and variation of historical events—essentially expanding a few hundred historical records into a set of tens of thousands of stochastic events. This approach was not the technically optimal solution, but rather only a feasible method under specific historical constraints of its time. However, this paradigm has remained fundamentally unchanged for over nearly four decades. Massive historical cumulative investments and the long-term accumulation of specialized knowledge have created profound path dependency and technological lock-in. In the era of AI, the opportunity cost of maintaining this path dependency has risen sharply.

This study identifies the following structural deficiencies inherent in the traditional SES methodology.

***(1) High Costs and Low Efficiency***

This is the most prominent issue. Model development is heavily dependent on multi-disciplinary teams, leading to protracted development cycles (often exceeding two years for a single major model) and prohibitive labor costs (reaching tens of millions of dollars). Hard-coded architectures fix core logic within the code, resulting in the continuous accumulation of technical debt and manual errors. Version updates often take years to complete, and parameter shifts can trigger significant volatility in analytical results, directly affecting transaction pricing and creating substantial so-called ‘model risk’ (Plata, Gupta and Azaele, 2020; Higuera Roa et al., 2025).

***(2) Fragmentation and Unmodelled Areas***

Traditional methodologies adopt a fragmented *Region+Peril* development strategy (e.g., independent models for North Atlantic+Hurricane, China+Typhoon, or Europe+Winter Storm, etc.). These models are developed and operated in silos. Consequently, manually specified correlation parameters fail to capture the intrinsic teleconnections across regions and perils (such as those driven by the El Niño-Southern Oscillation), making it difficult to accurately assess losses from correlated or compound events. Furthermore, vast geographical areas remain unmodeled due to such constraints, perpetuating the protection gap in regions where risk transfer is most needed.

***(3) Lack of Temporal Resolution***

Traditional SES are essentially event snapshots that record only peak intensity and associated losses. They cannot reconstruct the full dynamic evolution of a hazard from genesis to decay on a daily basis. This limitation makes models inadequate for complex risks where vulnerability is time-sensitive (e.g., different growth cycles of crops or varying construction phases of infrastructure). The absence of temporal resolution also restricts the ability to assess restoration periods for Business Interruption (BI) insurance.

In summary, the traditional technical route of catastrophe modeling remains a historical legacy of 1990s computing constraints. In the age of AI and HPC, the opportunity cost of defending the traditional paradigm has exceeded the cost of transformation, and the commercial imperative for change has never been stronger.

### 1.3 Advancements in AI for the Meteorological Sector

In recent years, AI has achieved breakthrough progress in the field of meteorology. Leading Large Meteorological Models (LMMs), such as the FourCastNet series developed by NVIDIA and Google DeepMind's GraphCast, have delivered an order-of-magnitude increase in computational efficiency. A single-node GPU can now complete a global 10-day weather forecast in minutes or even seconds, with the majority of evaluation metrics outperforming the conventional Numerical Weather

Prediction (NWP) models of the European Centre for Medium-Range Weather Forecasts (ECMWF). Furthermore, the precision of AI models in predicting critical hazard indicators—including tropical cyclone tracks, extreme temperatures, and atmospheric rivers—has begun to rival or even surpass traditional numerical methods (Radford, Ebert-Uphoff and Stewart, 2025). This precision is vital for capturing tail events in catastrophe risk assessment (Lam et al., 2022; Kurth et al., 2023; Shu et al., 2024). Other notable contributions include the Pangu-Weather model and various domain-specific LMMs tailored for specialized weather phenomena.

Beyond forecasting, AI downscaling technology has also reached significant milestones. By leveraging deep learning to extract cross-scale correlations, these models efficiently transform low-resolution data into kilometer-scale high-resolution output. Mardani et al. proposed a Residual Corrective Diffusion Modeling framework that achieves kilometer-scale atmospheric downscaling while effectively capturing the stochastic nature of multi-scale meteorological features (Mardani et al., 2024; 2025). Similarly, Tomasi et al. applied Latent Diffusion Models to dynamical downscaling, validating AI's potential to substitute traditional NWP methods (Tomasi et al., 2025; Tomasi, Franch and Cristoforetti, 2025). Rampal et al. utilized Generative Adversarial Networks (GANs) to enhance the statistical properties of climate downscaling, offering new perspectives for the reconstruction of extreme weather events (Rampal et al., 2022;

2025). These advancements mitigate the non-physical anomalies often associated with pure data-driven approaches, laying the foundation for relevant application in AI-based CAT modeling

The insurance industry has began to explore the integration of these AI technologies into CAT risk assessment. A notable example is AXA's utilization of the NVIDIA Earth-2 platform to generate ensemble reconstructions of historical hurricane scenarios, such as Hurricane Milton and Helene (NVIDIA, 2026). However, a systematic theoretical framework and an industry-grade solution has yet to emerge.

Crucially, the widespread adoption of open-source licensing for AI-related technologies provides essential infrastructure for a paradigm shift in CAT modeling. AI meteorological models, algorithms, and pre-trained model weights are predominantly released under permissive licenses, such as the MIT License or Apache License 2.0, and publicly accessible via on-line repositories such as GitHub. This openness provides the technical basis to dismantle traditional proprietary intellectual property barriers and catalyze distributed innovation in catastrophe risk quantification.

## 1.4 Research Scope and Contributions

The primary objective of this study is to propose a novel, AI-driven framework for SES generation and to provide initial empirical evidence for the feasibility and cost-efficiency potential of its core mechanism. The research scope is strictly defined as follows:

***Scope 1: TheTAISE Framework***

Drawing on the concepts of digital twins and meteorological ensemble forecasting, this study proposes the TAISE Framework designed to generate hazard events in a manner that mimics natural physical processes. The TAISE is designed to capture cross-regional teleconnections at global scale and the dynamic temporal evolution of disasters. It comprises of three sub-modules: (1) the Genesis Module (SES Prototyping), which generates continuous atmospheric sequences; (2) the Refinement Module (SES Downscaling and Intensity Calibration), which enhances spatial precision and corrects extreme value biases; and (3) the Ensemble Calibration Module (SES Statistical Alignment), which Conducts overall verification of the refined event samples and align SES with climatological norms.

***Scope 2: Preliminary Feasibility Testing of the Genesis Module***

As the foundational component of the TAISE Framework, the Genesis Module utilizes the generative capabilities of AI weather forecasting models to produce long-term, continuous atmospheric sequences. This study focuses on validating the logical consistency of this self-iterative approach. It is critical to note that the output of Genesis serves as 'event prototypes' or 'raw seeds', i.e. scenarios that require subsequent refinement and calibration before actuarial application.

***Scope 3: Performance and Efficiency Analysis***

The study evaluates the implementation efficiency of the Genesis Module, with particular focus on its computational cost structure relative to traditional CAT SES modeling methodologies.

The study is strictly scoped to Technology Readiness Level (TRL) 3-4. The following elements are identified as subsequent research phase, and are excluded explicitly in the study presented from the present work:

(1) Long-cycle simulation: The empirical testing in this study is strictly delimited to a certain simulatin period (50 years), which is sufficient to validate the logical stability and computational efficiency of the self-iterative mechanism. Extending simulations to millennial scales or conducting exhaustive verification of long-term physical consistency falls within the domain of meteorological research, and is not a prerequisites for establishing the technical feasibility and cost-structure advantages of the proposed framework.

(2) Actuarial closed-loop validation: Comprehensive validation of the Refinement and Ensemble Calibration Modules falls outside the scope of this initial proof-of-concept. These modules, essential for transforming

'raw SES', into 'refined SES', are identified as subsequent engineering phases.

(3) Commercial benchmarking: Constructing, calibrating, and benchmarking a production-ready system against established commercial models is not within this research's scope. Such an undertaking would contradict the objective laws of technological evolution and exceeds the mandate of academic exploratory research.

## 2. Methodology

### 2.1 Theoretical Foundation

The primary methodological innovation of this study is the 'functional repurposing' of AI weather forecasting models. Specifically, we propose decoupling their forecasting function to transform them into physically consistent scenario generators. Large-scale AI weather forecasting models (e.g., FourCastNet, GraphCast) were originally designed for medium-term weather forecasting, with an effective predictive window typically spanning 10 to 15 days. Beyond this horizon, predictive reliability diminishes rapidly due to the chaotic nature of atmospheric systems (Lam et al., 2022; Pathak et al., 2022; Kurth et al., 2023; Guo et al., 2024). However, this limitation in predictive accuracy does not constrain the construction of Stochastic Event Sets (SES).

Trained on vast historical data, AI weather forecasting models have learned to approximate the fundamental principles of atmospheric dynamics. Consequently, they are capable of generating weather fields (encompassing all observational elements such as temperature, pressure, solar radiation, humidity, precipitation, wind etc.) that implicitly adhere to natural physical constraints and maintain logically sound inter-variable relationships.

The TAISE framework leverages this capability to construct the SES. Its core mechanism is self-iterative scenario generation. Starting with any historical weather field as the initial input, the model outputs a physically consistent field for the subsequent $m$ days. By utilizing the output of day $m$ as the input for the next cycle, the framework iteratively constructs ultra-long-term, continuous atmospheric sequences. While this process is decoupled from real-time temporal anchoring, it preserves the climatological dimension—including seasonal cycles and daily-evolving atmospheric conditions. This approach systematically explores the space of potentially plausible physical scenarios, thereby satisfying the Law of Large Numbers essential for catastrophe risk analysis.

Beyond an apporach of computational brute force, such an exhaustive strategy is a probabilistic exploration of the potential state space within a complex system. The methodology shares a common lineage with recent

AI breakthroughs in fields such as structural biology and generative iconography, where generative models are used to map the possibilities of complex, high-dimensional spaces (reference).

## 2.2 Overview of the TAISE Framework

The TAISE framework adopts a three-stage pipeline architecture, The strategy can be summarized as 'High-Efficiency Generation followed by Low-Cost Calibration'.

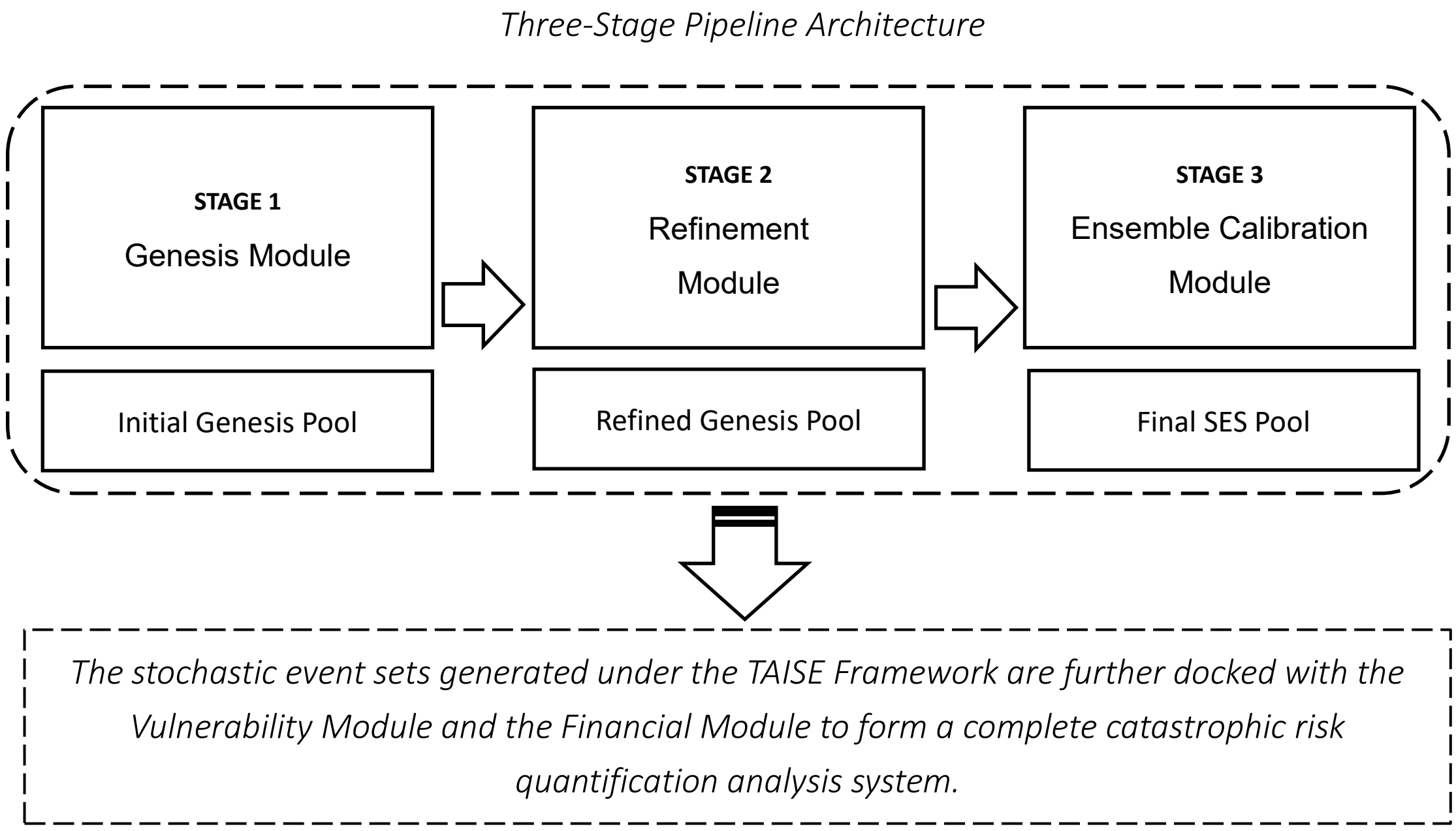

**Figure1:** Schematic Diagram of the Three-Module Pipeline Architecture of the TAISE Framework

### *Stage 1: Genesis Module (Stochastic Event Prototyping)*

In this framework, a ***Genesis*** is defined as a stochastic event prototype, which is a raw sample possessing the fundamental characteristics of an extreme weather event, serving as the basis for subsequent refinement. The primary objective of this module is to leverage AI models and a cyclic iterative generation logic to construct a vast pool of ‘genesis’, effectively creating the initial raw database for the SES.

***Stage 2: Refinement Module (Event Downscaling and Intensity Calibration)***

AI meteorological models often produce outputs with limited spatial resolution (e.g., 0.25 degrees) and may suffer from the 'smoothing' of extreme values, leading to an underrepresentation of peak intensities. The Refinement Module receives the low-resolution meteorological field output from the Genesis Module. It applies deep learning-based AI downscaling technologies, such as those outlined in section 1.3, to increase spatial resolution to the target scale while maintaining physical consistency across all variables.

To address the potential systematic underestimation of extreme event intensities, the module performs optimization and adjustment using algorithms based on regional disaster characteristics, such as Extreme Value Theory (EVT), Generalized Pareto Distribution (GPD), etc. Calibration is further conducted using multi-source information, such as historical observation data and insurance operational records. This ensures that the intensity-frequency distribution of the generated events remains consistent with the historical climatic experience of the target region. By refining the Genesis prototypes, this stage transforms the raw sample pool into a high-fidelity SES candidate set .

***Stage 3: Ensemble Calibration Module***

This module performs comprehensive statistical verification on the refined event samples to ensure that the overall statistical characteristics of the sample pool align with the historical climatic background of the target region. This includes cross-checks for temporal continuity, spatial distribution, density, and intensity frequency distributions. The result is a quality-assured SES, ready for integration with vulnerability and financial modules.

**2.3 Detail Implementation of The Genesis Module**

The Genesis module utilizes a self-iterative strategy to generate continuous atmospheric fields. The process begins by acquiring meteorological field data that fulfills the specific input requirements of the selected meteorological forecast model, for instance, global weather field data of any given date derived from ERA5 reanalysis data. This data serves as the initial state or 'seed' for the generation process, and denoted as *Day 0*. The model then outputs the global atmospheric fields for the subsequent $m$ days, e.g. *Day 1* to *Day m* in the *1st* iteration. Following this, the iterative process commences, utilizing the global weather field of the final day from the previous output as the input for the next computational cycle.

The technical workflow is defined as follows:

Let the iterative step-length be $m$ days, and the global daily atmospheric field be represented by $W$.

The input for the $i$-$th$ iteration of model computation is the global weather field from day $(i-1)m$:

$$Input_i = W_{(i-1)m}$$

The $i$-$th$ iteration then generates a sequence of weather fields for the following $m$ days:

$$Output_i = \{W_{(i-1)m+1},\ W_{(i-1)m+2},\ \ldots,\ W_{im}\}$$

By utilizing the last day of output $Output_i$, i.e. ,the $im$-$th$ day $W_{im}$ as the initial input for the $(i+1)$-$th$ iteration, a continuous iterative chain is established. Through this self-iterative mechanism, by utilizing the atmospheric dynamic constraints learned by the AI model, ultra-long daily meteorological sequences meeting physical rationality can be continuously generated, thus realizing 'physical emergence' rather than 'statistical construction'.

### 2.4 Genesis Module Experiment Design

#### 2.4.1 Experiment Boundaries and Scope

As noted in section 1.4, this experiment serves as a preliminary feasibility study, categorized at Technology Readiness Level (TRL) 3-4. Its core contribution is the empirical demonstration that the self-iterative mechanism of AI meteorological models can efficiently generate continuous atmospheric fields. A comprehensive validation—including full physical consistency across multi-variable meteorological dynamics and millennial-scale simulations—is designated for subsequent research stages (TRL 6-7).

The core boundaries of this experiment are defined as follows. The selection of a 50-year period is emphasized, as a longer simulation is unnecessary for this initial feasibility test (e.g., longer runs would incur computational costs without providing incremental insight into the core iterative mechanism). The spatial focus on North American was selected not only for data storage optimization but also because it is a core insurance market. The choice of cyclones allows for a focused test of the extreme event detection algorithm without the complexity of multi-peril data density, which would otherwise distract from the feasibility objective.

| **Dimension** | **Experimental Scope** | **Rationale / Specification** |
|---|---|---|
| Temporal Scale | 50 Years (18,250 days) | Validating mid-term iterative stability to provide a technical foundation for millennial-scale simulations. |
| Spatial Scope | Primary focus on North America | Focusing on a core insurance market region to optimize data storage and analytical focus. |
| Peril Scope | Tropical and Extratropical Cyclones | Prioritizing one of the highest contributors to global insured losses to manage data density. |
| Depth of Validation | Efficiency metrics and extreme event generation capability | Excludes quantitative benchmarking against legacy models or in-depth atmospheric physics mechanism analysis. |

**Table 1:** Genesis Module Experimental Boundaries

### 2.4.2 Model and Data

The experiment utilizes FourCastNet V2.0, a global data-driven high-resolution weather forecasting model developed by NVIDIA. Built upon the Adaptive Fourier Neural Operator (AFNO), FourCastNet was trained on ERA5 reanalysis data and covers 20 critical meteorological variables (including temperature, wind fields, pressure, and specific humidity). It generates medium-term forecasts at a 0.25° resolution and supports rapid inference on a single GPU. The model accepts a global weather field at a single timestamp as input and outputs a 10-day global forecast. Due to its architecture based on spherical Fourier transforms, the model must output the complete global field.It does not support regional subsetting or variable subsetting during the inference stage. The model is released under the Apache 2.0 License and is publicly accessible via GitHub.

ERA5 Reanalysis Data is employed as the initial input for the iterative process. Produced by the European Centre for Medium-Range Weather Forecasts (ECMWF), ERA5 provides global meteorological reanalysis from 1940 to the present. It is open-source under the Copernicus License, available for public download via the Copernicus Climate Change Service (C3S) Climate Data Store (CDS).

Inference computations for this experiment were conducted on a single NVIDIA V100 GPU.

#### 2.4.3 Data Filtering and Storage Strategy

The volume of global weather field data is immense. A single day of 0.25° resolution data contains approximately 2.6 million grid points, with multi-parameter storage requirements reaching gigabyte levels per day. Direct storage of long-term sequences would exert petabyte-level (PB) storage pressure. Consequently, a feature-driven selective storage strategy was implemented:

**(1) *Geographical Subsetting***

The experiment extracts data only for the North American region (25°N–60°N, 170°W–50°W), comprising 67,200 grid points. Data for all other regions is discarded immediately after generation.

***(2) Real-time Feature Identification***

A cyclone detection algorithm is applied to the generated daily weather fields within the specified geographical scope to identify potential extreme weather events.

***(3) Selective Archiving***

Only weather fields that fall within the specified region and contain identified cyclone features are retained. Redundant data (outside the region or within the region but lacking specific features) is deleted.

### 2.4.4 Experimental Objectives

The specific objectives of the experiment are:

***Objective 1:*** To evaluate the generation efficiency of the Genesis Module, including computational time and data storage optimization.

***Objective 2:*** To verify the stability of the Genesis Module in generating extreme weather events.

***Objective 3:*** To assess the production cost structure of the Genesis Module.

## 3. Experimental Results

### 3.1 Computational Efficiency

The experiment generated a 50-year (18,250-day) global daily atmospheric sequence through 1,825 iterative cycles. As shown in Table 2, under a single V100 configuration, the average duration for a single iteration (generating 10 days of global weather fields) was 128 seconds. Within this cycle, model inference accounted for 80% of the time (102.0s ± 8.0s), while data processing tasks, such as I/O, accounted for the remaining 20% (approximately 26 seconds). This temporal structure reveals two critical characteristics. First, the dominance of inference time (80%) indicates that the computation is concentrated within the FourCastNet operations rather than data movement or pre-processing. This suggests that upgrading GPU hardware (e.g., transitioning to A100/H100 GPUs or multi-GPU arrays) could yield near-linear performance gains. Second, the overhead for I/O and data processing remained relatively low (20%), with raw data I/O taking 18 seconds (14%) and the extraction of the target region (North America) requiring 8 seconds (6%). The latter depends on specific experimental strategies and may vary slightly on different experimental strategies.

Throughout the experimental cycle, model inference time remained highly stable, with a mean of 102 seconds per 10-day block and a P95 latency of 115.4 seconds (a deviation of less than 15% from the mean). Such consistency is vital for the cost predictability of millennial-scale simulations.

The experiment achieved a sustained output of 18.49 simulation-years per calendar day (with an hourly equivalent throughput of 281.2 simulation-days per computational hour). The total computation time required to complete the 50-year simulation was 64.9 hours (approximately 2.7 days).

| Metric | Value | Description |
|---|---|---|
| Total Simulation Period | 50 years | 18250 days |
| Iteration Strategy | 10-day chunks | 1825 iterations |
| Mean Iteration Time | 128.0s | ~2.1minutes |
| Model Inference Time | 102.0s±8.0s | 80% of total |
| I/O Overhead Data processing | 26.0s | Including subset slicing |
| Total Computation Time | 64.9 Hours | ~2.7 days |
| Throughput | 281.2 days/hour | 18.49 simulated year per calendar day |
| P95 Latency | 115.4s | 95th percentile inference time |

**Table 2:** Computational Efficiency Breakdown of the Genesis Module

## 3.2 Capacity for Generating Extreme Weather Conditions

The experiment focused on the North American region (25°N–60°N, 170°W–50°W). Utilizing a cyclone detection algorithm based on wind field structures, a total of 2,157 independent cyclonic events were identified over the 50-year simulation period. Detected cyclones exhibited distinct counter-clockwise rotation, aligning with the laws of Northern Hemisphere atmospheric motion driven by the Coriolis effect.

The generated cyclones displayed characteristic wind speed minimum zones (such as the 'eye of the storm' in tropical cyclones or the low-pressure center in extratropical cyclones), with spatial scale of the cyclone systems predominantly concentrated between 1,000 and 3,000 kilometers. These dimensions are basically consistent with the actual climatological scale characteristics of mid-latitude extratropical, tropical cyclones including explosive cyclogenesis. Spatially, the activity was concentrated across the North American landmass and its adjacent maritime regions—including the Gulf of Alaska, U.S. and Canadian west coasts and the Western Atlantic—matching observed core activity corridors for North American cyclones. Throughout the entire simulation process, the framework demonstrated high stability, with no evidence of systemic data gaps or physical incongruities.

| **Feature** | **Genesis Output** | **Physical Meaning / Real-world Reference** |
|---|---|---|
| Rotation Direciton | Counter-ClockWise | Consistent with the Coriolis effect in the Northern Hemisphere |
| Windspeed Minimum Zone | Present | Typical structure of a cyclone |
| Windspeed Distribution | Asymmetric | Typical structure of a cyclone |
| Spatial Scale | 1000+km Scale | 1000+km Scale |
| Frequency | ~40 per Year | Related to cyclone statistical thresholds: up to 100 per year if low-intensity cyclones are included; |

| | 25–35 large-scale cyclones per year (including tropical cyclones, extratropical cyclones, from both Pacific and Atlantic basins, etc. (reference); |
|---|---|

**Table 3：** Detection of potential extreme conditions in the Genesis Module experiment

.

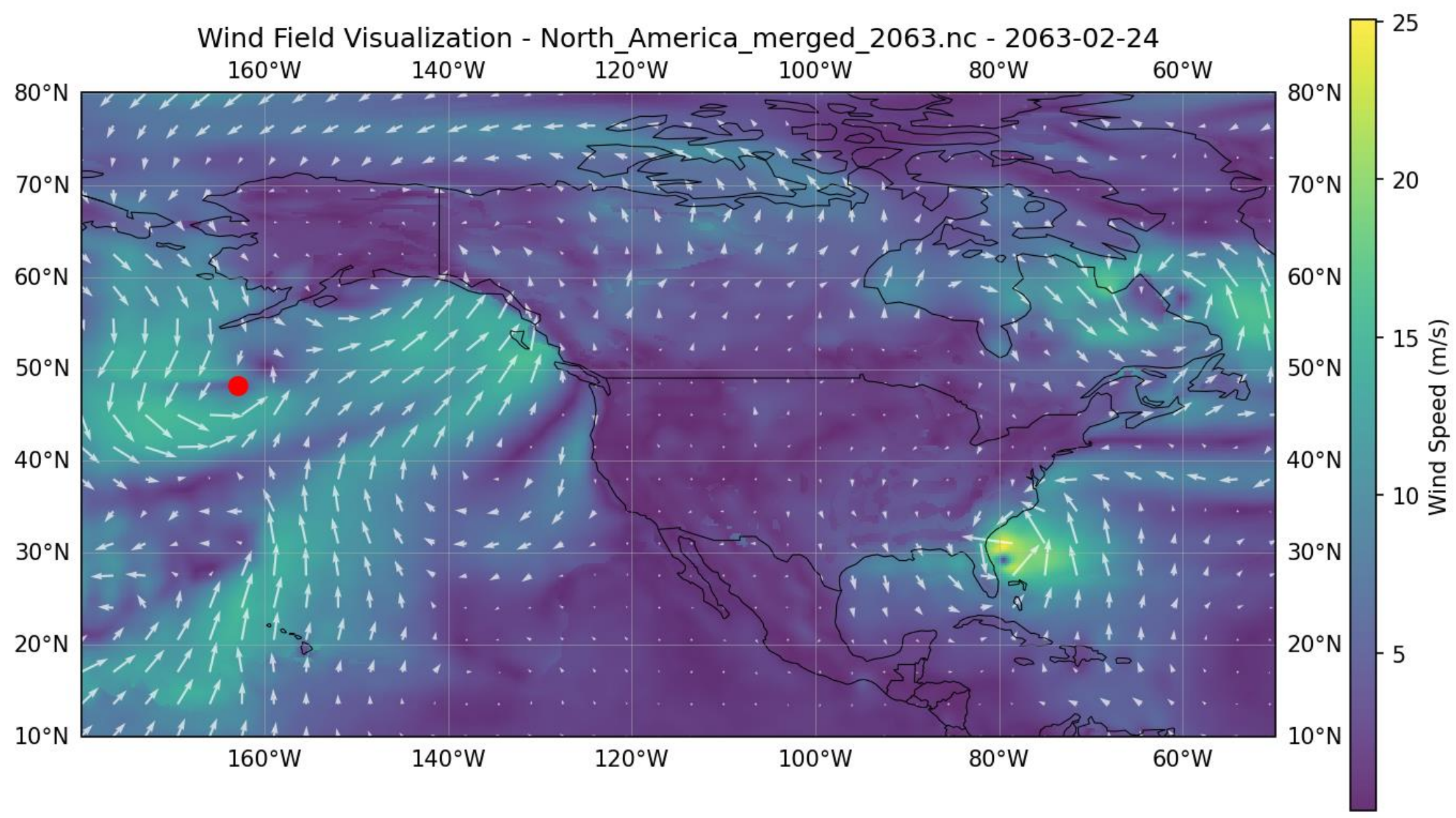

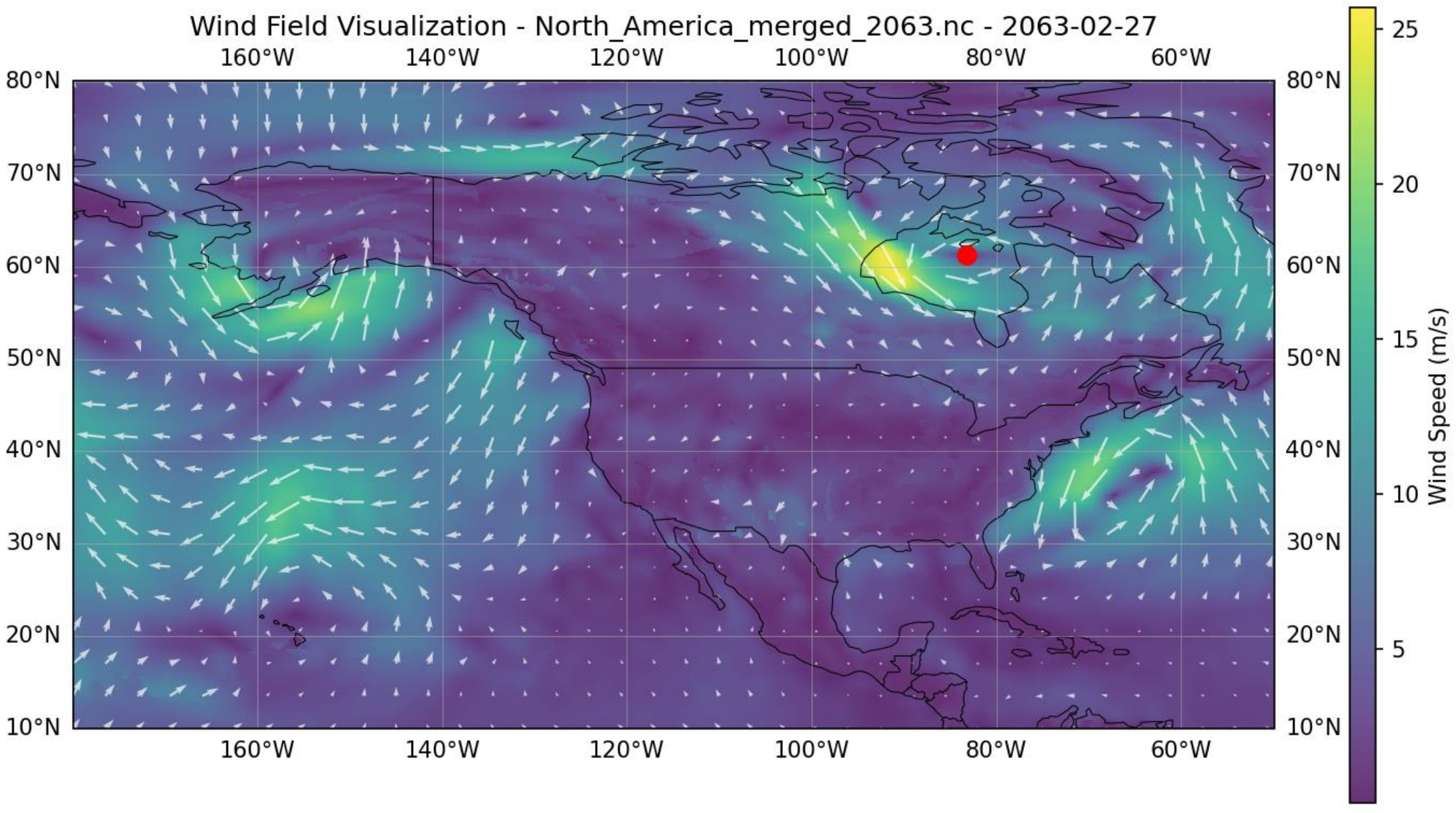

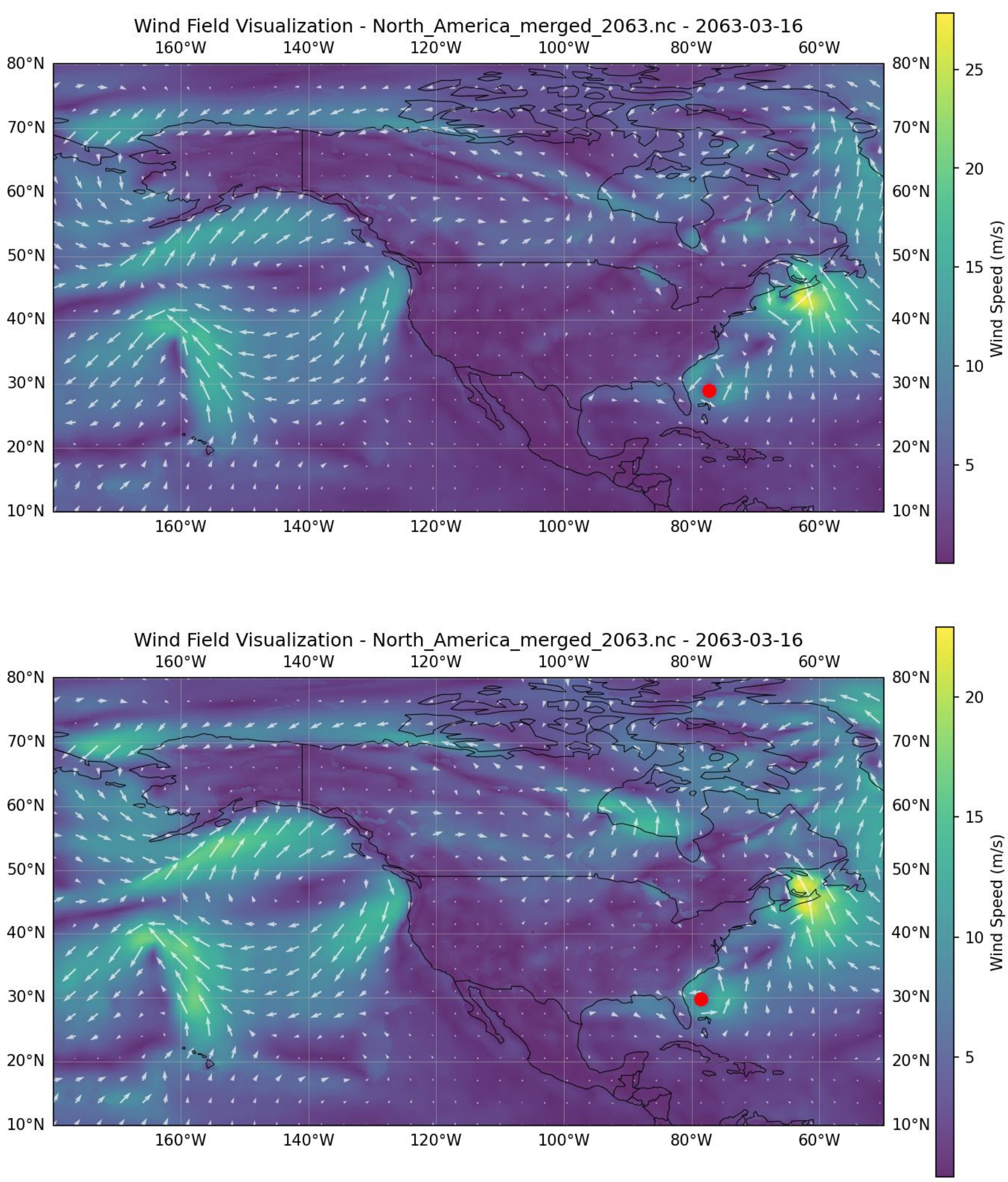

**Figure 2（a-d）**Visualization of North American cyclones with red dots marking cyclone eyes. Arrows indicate wind directions, with their size proportional to wind speed. The timestamp on the map, such as 2063-03-16 is used as a schematic year for model configuration and computational convenience, rather than a real-world projection as explained in section 2.1.

This observed stability stems from FourCastNet's learned atmospheric dynamical constraints, providing a robust foundation that sufficiently supports the feasibility validation objectives of this research. While the intensity of the detected cyclones exhibits an underestimation relative to historical observations, this is a recognized limitation inherent in ERA5-derived training data. This systemic intensity bias is addressed in Section 4.3.3 as a calibration challenge for the subsequent Refinement module. As indicated in experiment boundary, a comprehensive meteorological statistical analysis remains outside the scope of the current experiment.

### 3.3 Cost Structure Analysis

The experiment utilized a single NVIDIA V100 GPU instance. Its usage market hourly rates vary with purchasing schemes including on-demand, 1-year saving plan or 3-year saving plan etc. For illustration purpose, this study adopts an approximate average value between the maximum and minimum rate within market range. For the above mentioned V100x1 instance, an approximate market rate is approximately $2/hour. The theoretical computational cost for the 50-year simulation was approximately $130 (based on 65 hours of computation, exclusive of peripheral costs such as debugging, focusing solely on iterative inference). Based on theoretical linear extrapolation, generating a 10,000-year global daily atmospheric sequence would require approximately 13,000 compute hours (roughly 540 calendar days on a single V100 GPU), resulting in an approximate linear estimation of computational cost of roughly $26,000. Compared to the V100, a commercial A100 GPU offers 2 to 2.5 times the inference performance. Utilizing a standard 8-chip A100 instance (e.g., AWS p4d.24xlarge instance) would provide over 20 times the throughput of the single V100 used in this study. This could reduce the simulation time for a 10,000-year sequence from 13,000 hours to roughly 650 hours. Assuming a market rate of $20/hour for AWS p4d.24xlarge instance, the total cost for 650 hours would be approximately $13,000—representing a 50% reduction in cost compared to a single V100 approach, alongside a 20-fold increase in speed. The potential for efficiency gains is even greater with H100 chips (per single chip, a H100 offering over 15 times the performance of a V100) or multi-node clusters.

Regarding data management, the North American region selected for this study represents only 6.5% of the Earth's surface. While the model originally outputs 20 meteorological parameters, this experiment retained only wind-speed-related data. Under this selective storage strategy, a single 10-day output (North America, wind speed) occupies 92 MB. Conversely, full global coverage of all variables would exceed 2 GB per 10-day block, accumulating to at least 730 TB for a 10,000-year sequence. At standard cloud storage rates (e.g., AWS S3 at $0.023/GB/month), annual storage costs could exceed $200,000. This highlights that the storage strategy chosen during development is a primary driver of long-term operational costs.

Traditional methodologies are generally associated with massive sunk costs. Leading vendors allocate hundreds of millions annually to R&D. Consequently, licensing fees are often structured as a percentage of a client's transaction volume; for instance, major international reinsurance brokers may pay upwards of $10 million annually in licensing fees to a single vendor.

In comparison, the Genesis Module demonstrates the potential for the capitalization of computational infrastructure. Through a relatively low one-time investment in compute power, it is possible to generate a global, multi-peril millennial-scale scenario library that supports infinite reuse and flexible adjustment. Although the Genesis Module is only one component of the TAISE framework—and this analysis does not include the development of Refinement, Calibration, Vulnerability, or Financial modules—its disruptive potential cost advantage is evidently significant.

| **Computation Instance (AWS)** | **V100x1 (p3.2xlarge)** | **A100x8 (p4de.24large)** | **H100x8 (p5.48xlarge)** |
|---|---|---|---|
| computation capability | ~125TFLOPS | ~2500TFLOPS | ~15800TFLOPS |
| Approximate hourly rate (avarge of different purchase plans) | ~2$/hour | ~20$/hour | ~45$/hour |
| 10,000-year Global Daily Fields (computation time consumption) | ~13,000 hours | ~650 hours | ~103 hours |
| Computation cost (linear theoretical estimation) | ~$26,000 | ~$13,000 | ~$4,635 |

**Table 4:** Illustrative Theoretical Cost Comparison - Computational Infrastructure Only (Excluding Validation, Calibration etc.)

**Note:** All figures for 10,000-year simulations are linear theoretical extrapolations based on the 50-year experiment. They assume perfect hardware scalability and do not account for (i) error accumulation mitigation costs; (ii) storage of multi-parameter global fields, or (iii) actuarial validation expenditures. These estimates are intended solely as indicative of raw compute economics, not as operational cost projections for insurance-grade SES production.

## 4. Analysis and Discussion

This study demonstrates a methodology for repurposing the core predictive capabilities of AI weather forecasting models into a foundational tool for CAT risk modeling. The proposed TAISE Framework facilitates a novel AI-based approach to SES generation. By introducing the Genesis Module's self-iterative mechanism and conducting experiments based on the FourCastNet model, this research has generated a 50-year global daily high-resolution atmospheric sequence, preliminarily validating the technical feasibility of the approach at Technology Readiness Level 3-4 (TRL 3-4).

### 4.1 Preliminary Observations on Computational Cost Structure

The high development, licensing, and maintenance costs associated with traditional commercial CAT models have created entry barriers, which are increasingly misaligned with the rapid advancement of AI capabilities and HPC infrastgructure.

Unlike traditional methods, where version updates, regional expansions, and the addition of new perils incur high marginal costs due to repetitive manual labour, the TAISE Framework reshapes the cost structure from high fixed/high marginal to low fixed/low marginal. The experiment utilized a single NVIDIA V100 GPU to generate a 50-year global daily sequence in an approximate of 65 hours. As AI performance continues to improve and cloud computing costs decline, this efficiency gap will continue to widen further. The removal of cost barriers could accelerate the restructuring of current market dynamics, shifting pricing models from high licensing fees toward value-added services. This democratization could enable small insurers, emerging-market risk pools, and public-sector disaster management agencies to access quantification infrastructures that were previously prohibitively expensive. Based on inudstry practice and the author's professional experience, in traditional CAT model development, SES construction typically represent at least 60% of total model development effort, driven primarily by labour intensity. Major CAT modeling firms with annual revenues exceeding $100 million USD (in some cases exceeding 300 million) typically allocate 50-60% of their revenue to R&D. Developing a mainstream CAT model often cost close or over $10 million. It is therefore reasonable to estimate that it is not uncommon for the cost of SES development under traditional appoarch to be at least $5 millon USD.

## 4.2 Potential Technical Advantages

### 4.2.1 Systemic Holism

The TAISE Framework constructs weather fields using a unified global logic, allowing extreme weather to emerge pseudo-naturally across the globe without the need for explicit hard-coded physical formulas. Events emerge in a pseudo-natural manner— their location, footprint, intensity, and frequency are all outputs of model inference, rather than prescribed parametric assumptions. This enables the framework to capture long-distance teleconnections and global-scale dependencies that traditional regionalized modelling approaches cannot accurately characterize. The scenarios generated reflect complete climatological cycles rather than isolated event snapshots, providing more scientifically grounded inputs for quantifying compound and cascading risks.

#### 4.2.2 Temporal Continuity

In contrast to the event-snapshot approach of traditional SES methodologies, the TAISE Framework introduces continuous temporal modeling to the CAT risk domain. By outputting daily sequences, it reconstructs the temporal evolution of weather systems from genesis through intensification, peak impact, and dissipation. This spatiotemporal continuity enables more precise assessment of time-sensitive exposures, such as Business Interruption insurance, supply chain disruption, and multi-event aggregate reinsurance covers, and lays the foundation for dynamic risk assessment and scenario-responsive pricing.

#### 4.2.3 Scalability and Extensibility

The TAISE Framework demonstrates high adaptability potential. Beyond actuarial pricing application for insurers, ultra-long daily continuous global weather field sequences provide a foundation to support broader financial industry's climate risk stress testing requirements and disaster management planning for public-sector agencies. This capability for cross-domain reuse significantly reduces marginal expansion costs. By adjusting initial conditions or integrating Genesis Module output with IPCC Shared Socioeconomic Pathway (SSP) scenarios, such as Coupled Model Intercomparison Project Phase 6 (CMIP6) projections, the TAISE Framework has the potential to reflect climate-driven baseline drifts, providing technical infrastructure for forward-looking risk assessment and climate-related financial disclosures.

### 4.3 Addressing Current Limitations

This study acknowledges technical limitations. As defined in the research scope (Section 1.4), these represent tractable engineering challenges rather than fundamental paradigm flaws. They can be mitigated through established low-cost methods implemented in the proposed Refinement and Ensemble Calibration modules.

#### 4.3.1 Error Accumulation

The self-iterative mode carries an inherent risk of error accumulation over extended time periods, a common challenge in the numerical simulation of chaotic systems. The 50-year experiment demonstrates preliminarily that within this timeframe, physical consistency remains stable. For millennial-scale simulations (10,000 years), we propose a segmented calibration and sequential generation strategy. By employing interval cycles (e.g. 5-year segments) and re-initialising the model with real observational data (e.g. ERA5 reanalysis) at each interval, the approach preserves core computational efficiency while ensuring long-term meteorological plausibility.

#### 4.3.2 Resolution Constraints

The native 0.25° resolution (approx. 31km) is coarser than the 1–5km resolution often required for local peril modeling. This limitation is designed to be addressed by the Refinement Module, which utilizes increasingly mature AI-based downscaling techniques to enhance spatial precision efficiently, such as those surveyed in section 1.3. Notably, this field is experiencing rapid technological progress, with continuous emergence of improved methods, models and training strategies.

#### 4.3.3 Underestimation of Extreme Intensities

Due to the smoothing effects inherent in the ERA5 training data, the intensity of extreme events may be underestimated, a well documented characteristic of data-assimilation products (Hersbach et al., 2020). The core function of the Genesis Module is to generate physically coherent event prototypes—capturing their temporal evolution, spatial footprint, and track dynamics. Intensity underestimation is a tractable calibration issue that can be corrected in the Refinement and Ensemble Calibration modules using established techniques such as Extreme Value Theory (EVT), quantile mapping, and historical benchmarking against observed extreme event statistics. This ensures that the final refined and calibrated SES is both physically plausible and actuarally robust.

It should be emphasised that intensity bias (underestimation or overestimation) is not unique to the TAISE Framework. Traditional approaches also requires extensive calibration and adjustment to align with empirical oss distributions and tail-risk characteristics, indicating that this issue is inherent to the CAT modelling domain rather than a limitation specific to the AI-based TAISE.

**5. Conclusion**

To address the structural challenges and limitations of CAT modeling, this study proposes a novel AI-based approach (TAISE framework), which shifts the SES generation from manually-led extrapolation to AI-driven pesudo-natural scenario emergence. The TAISE Framework achieves this by repurposing deep learning meteorological models as physically self-consistent weather field generators. Through a self-iterative mechanism, extreme events emerge pseudo-naturally based on the inherent principles of atmospheric dynamics. The 50-year global simulation conducted validates the technical feasibility of this approach at a preliminary level. Utilizing a single NVIDIA V100 GPU, the experiment achieved a throughput of 18.49 simulation-years per calendar day, stably generating 2,157 physically reasonable cyclonic events. The theoretical linear projection of computational cost for a millennial-scale scenario library is estimated at under $5,000 (Table 4)—representing an order-of-magnitude cost reduction compared to traditional modeling technical route.

The TAISE Framework is not presented as a perfect solution. Its core value lies in demonstrating a radical AI-based cost-efficiency asymmetry, i.e. the potential to achieve fidelity levels that meet or exceed current industry benchmarks at a fraction of the resource expenditure. This AI-based asymmetry in cost-effectiveness could fundamentally alter the economics of CAT risk quantification, with direct implications for insurance market, and the feasibility of covering protection gaps in underserved areas. The study acknowledges several limitations. However, these are considered foreseeable technical debts inherent in the progression of technology readiness, rather than fundamental flaws in the paradigm itself. The study presented in this paper should be viewed against the backdrop of AI's extremely rapid development trajectory, not only in meteorology models, but also in Large Language Models (LLMs), AI coding assistants, and automated workflow tools etc. With continued breakthroughs anticipated over the next 3-5 years, the labor costs associated with CAT modeling will be further compressed, leading to a radical transformation of the current commercial ecosystem. This study explore a strategic roadmap for the transition of CAT modeling from the PC era to the era of AI.Future research directions should prioritize the following, including but not limited: (1) Conducting longer period simulations to test the effectiveness of segmented calibration strategies and meteorological robustness. (2) Validation of Refinement and Ensemble Calibration Modules, and Interfacing the framework with actuarial pricing and regulatory compliance requirements. (3) Adapting the framework to IPCC SSP dynamic baselines to address climate-driven risk shifts.